\UseRawInputEncoding
\documentclass[11pt,a4paper]{article}
\usepackage[hyperref]{eacl2021}
\usepackage{times}
\usepackage{latexsym}
\usepackage{subcaption}
\usepackage{comment}
\usepackage{enumitem}
\usepackage{multirow}
\usepackage{booktabs}
\usepackage{mathtools, nccmath}

\usepackage{microtype}

\aclfinalcopy 

\title{Attention Can Reflect Syntactic Structure \\\emph{(If You Let It)}}
\author{\thanks{~~Equal contribution. Order was decided by a coin toss.}\, Vinit Ravishankar\footnotemark[2]\quad \footnotemark[1]\, Artur Kulmizev\footnotemark[3]\quad Mostafa Abdou\footnotemark[4]\\ \textbf{Anders S{\o}gaard}\footnotemark[4]\qquad \textbf{Joakim Nivre}\footnotemark[3] \vspace{0.3em} \\
  \footnotemark[2]\; Language Technology Group, Department of Informatics, University of Oslo \\
  \footnotemark[3]\; Department of Linguistics and Philology, Uppsala University\\
  \footnotemark[4]\; Department of Computer Science, University of Copenhagen \hspace{2mm} \vspace{0.3em} \\ 
  \footnotemark[2]{~~\tt vinitr@ifi.uio.no} \\
  \footnotemark[3]{~~\tt \{artur.kulmizev,joakim.nivre\}@lingfil.uu.se} \\
  \footnotemark[4]{~~\tt \{abdou,soegaard\}@di.ku.dk}
}

\date{}

\begin{document}
\maketitle
\begin{abstract}
Since the popularization of the Transformer as a general-purpose feature encoder for NLP, many studies have attempted to decode linguistic structure from its novel multi-head attention mechanism. However, much of such work focused almost exclusively on English --- a language with rigid word order and a lack of inflectional morphology. In this study, we present decoding experiments for multilingual BERT across 18 languages in order to test the generalizability of the claim that dependency syntax is reflected in attention patterns. We show that full trees can be decoded above baseline accuracy from single attention heads, and that individual relations are often tracked by the same heads across languages. Furthermore, in an attempt to address recent debates about the status of attention as an explanatory mechanism, we experiment with fine-tuning mBERT on a supervised parsing objective while freezing different series of parameters. Interestingly, in steering the objective to learn explicit linguistic structure, we find much of the same structure represented in the resulting attention patterns, with interesting differences with respect to which parameters are frozen. 

\end{abstract}

\section{Introduction}

In recent years, the attention mechanism proposed by \citet{bahdanau2014neural} has become an indispensable component of many NLP systems. 
Its widespread adoption was, in part, heralded by the introduction of the Transformer architecture \citep{vaswani2017attention}, which constrains a soft alignment to be learned across discrete states in the input (self-attention), rather than across input and output \citep[e.g.,][]{xu2015show,rocktaschel2015reasoning}. The Transformer has, by now, supplanted the popular LSTM \citep{hochreiter1997long} as NLP's feature-encoder-of-choice, largely due to its compatibility with parallelized training regimes and ability to handle long-distance dependencies. 

Certainly, the nature of attention as a distribution over tokens lends itself to a straightforward interpretation of a model's inner workings. \citet{bahdanau2014neural} illustrate this nicely in the context of \texttt{seq2seq} machine translation, showing that the attention learned by their models reflects expected cross-lingual idiosyncrasies between English and French, e.g., concerning word order. With self-attentive Transformers, interpretation becomes slightly more difficult, as attention is distributed across words within the input itself. This is further compounded by the use of multiple layers and heads, each combination of which yields its own alignment, representing a different (possibly redundant) view of the data. Given the similarity of such attention matrices to the score matrices employed in arc-factored dependency parsing \citep{mcdonald05acl,mcdonald05emnlp}, a salient question concerning interpretability becomes: Can we expect some combination of these parameters to capture linguistic structure in the form of a dependency tree, especially if the model performs well on NLP tasks? If not, can we relax the expectation and examine the extent to which subcomponents of the linguistic structure, such as subject-verb relations, are represented? This prospect was first posed by \citet{raganato2018analysis} for MT encoders, and later explored by \citet{clark-etal-2019-bert} for BERT. Ultimately, the consensus of these and other studies \citep{voita2019analyzing,htut2019attention,limisiewicz2020universal} was that, while there appears to exist no ``generalist'' head responsible for extracting full dependency structures, standalone heads often specialize in capturing individual grammatical relations. 

Unfortunately, most of such studies focused their experiments entirely on English, which is typologically favored to succeed in such scenarios due to its rigid word order and lack of inflectional morphology. It remains to be seen whether the attention patterns of such models can capture structural features across typologically diverse languages, or if the reported experiments on English are a misrepresentation of local positional heuristics as such. Furthermore, though previous work has investigated how attention patterns might change after fine-tuning on different tasks \citep{htut2019attention}, a recent debate about attention as an explanatory mechanism \citep{jain2019attention,wiegreffe2019attention} has cast the entire enterprise in doubt. Indeed, it remains to be seen whether fine-tuning on an explicit structured prediction task, e.g. dependency parsing, can force attention to represent the structure being learned, or if the patterns observed in pretrained models are not altered in any meaningful way. 

To address these issues, we investigate the prospect of extracting linguistic structure from the attention weights of multilingual Transformer-based language models. In light of the surveyed literature, our research questions are 
as follows:

\begin{enumerate}[itemsep=0pt,topsep=3pt]
\setlength{\parskip}{.1mm}
\setlength{\parsep}{0.2mm}

    \item Can we decode dependency trees for some languages better than others?
    \item Do the same layer--head combinations track the same 
    relations across languages?
    \item How do attention patterns change after fine-tuning with explicit syntactic annotation?
    \item Which components of the model are involved in these changes?
\end{enumerate}
In answering these questions, we believe we can shed further light on the (cross-)linguistic properties of Transformer-based language models, as well as address the question of attention patterns being a reliable representation of linguistic structure. 


\section{Attention as Structure}

\paragraph{Transformers}
\label{para:transformers}

The focus of the present study is mBERT, a multilingual variant of the exceedingly popular language model \citep{devlin2019bert}. BERT is built upon the Transformer architecture \citep{vaswani_attention_2017}, which is a self-attention-based encoder-decoder model (though only the encoder is relevant to our purposes). A Transformer takes a sequence of vectors $\mathbf{x} = [\mathbf{x_1}, \mathbf{x_2}, ... \mathbf{x_n}]$ as input and applies a positional encoding to them, in order to retain the order of words in a sentence. These inputs are then transformed into query ($Q$), key ($K$), and value ($V$) vectors via three separate linear transformations and passed to an attention mechanism. A single attention head computes scaled dot-product attention between $K$ and $Q$, outputting a weighted sum of $V$:
\useshortskip \begin{equation}
\mathrm{Attention}(Q, K, V) = \mathrm{softmax}\left(\frac{QK^\top}{\sqrt{d_{k}}}\right)V
\end{equation}

\noindent
For multihead attention (MHA), the same process is repeated for $k$ heads, allowing the model to jointly attend to information from different representation subspaces at different positions \citep{vaswani_attention_2017}. Ultimately, the output of all heads is concatenated and passed through a linear projection $W^O$:
\useshortskip \begin{equation}
H_i = \mathrm{Attention}\left(QW_{i}^{Q},KW_{i}^{K},VW_{i}^{V}\right)
\end{equation}
\useshortskip \begin{equation}
\mathrm{MHA}(Q, K, V) = \mathrm{concat}(H_1, H_2, ..., H_k)W^O
\end{equation}

\noindent
Every layer also consists of a feed-forward network ($\mathrm{FFN}$), consisting of two Dense layers with ReLU activation functions. For each layer, therefore, the output of $\mathrm{MHA}$ is passed through a LayerNorm with residual connections, passed through $\mathrm{FFN}$, and then through another LayerNorm with residual connections.




\paragraph{Searching for structure}

Often, the line of inquiry regarding interpretability in NLP has been concerned with extracting and analyzing linguistic information from neural network models of language \citep{belinkov2019analysis}. Recently, such investigations have targeted Transformer models \cite{hewitt2019structural, rosa2019inducing, tenney2019bert}, at least in part because the self-attention mechanism employed by these models offers a possible window into their inner workings. With large-scale machine translation and language models being openly distributed for experimentation, several researchers have wondered if self-attention is capable of representing syntactic structure, despite not being trained with any overt parsing objective.

In pursuit of this question, \citet{raganato2018analysis} applied a maximum-spanning-tree algorithm over the attention weights of several trained MT models, comparing them with gold trees from Universal Dependencies \citep{nivre16lrec,nivre20lrec}. They found that, while the accuracy was not comparable to that of a supervised parser, it was nonetheless higher than several strong baselines, implying that some structure was consistently represented. \citet{clark-etal-2019-bert} corroborated the same findings for BERT when decoding full trees, but observed that individual dependency relations were often tracked by specialized heads and were decodable with much higher accuracy than some fixed-offset baselines. Concurrently, \citet{voita2019analyzing} made a similar observation about heads specializing in specific dependency relations, proposing a coarse taxonomy of head attention functions: \textit{positional}, where heads attend to adjacent tokens; \textit{syntactic}, where heads attend to specific syntactic relations; and \textit{rare words}, where heads point to the least frequent tokens in the sentence. 
\citet{htut2019attention} followed \citet{raganato2018analysis} in decoding dependency trees from BERT-based models, finding that fine-tuning on two classification tasks did not produce syntactically plausible attention patterns. Lastly, \citet{limisiewicz2020universal} modified UD annotation to better represent attention patterns and introduced a supervised head-ensembling method for consolidating shared syntactic information across heads.

\paragraph{Does attention have explanatory value?} \hspace{-2.5mm}Though many studies have yielded insight about how attention behaves in a variety of models, the question of whether it can be seen as a ``faithful'' explanation of model predictions has been subject to much recent debate. For example, \citet{jain2019attention} present compelling arguments that attention does not offer a faithful explanation of predictions. Primarily, they demonstrate that there is little correlation between standard feature importance measures and attention weights. Furthermore, they contend that there exist \textit{counterfactual} attention distributions, which are substantially different from learned attention weights but that do not alter a model's predictions. Using a similar methodology, \citet{serrano2019attention} corroborate that attention does not provide an adequate account of an input component's importance. 

In response to these findings, \citet{wiegreffe2019attention} question the assumptions underlying such claims. 
Attention, they argue, is not a \textit{primitive}, i.e., it cannot be detached from the rest of a model's components as is done in the experiments of \citet{jain2019attention}. They propose a set of four analyses to test whether a given model's attention mechanism can provide meaningful explanation and demonstrate that the alternative attention distributions found via adversarial training methods do, in fact, perform poorly compared to standard attention mechanisms. On a theoretical level, they argue that, although attention weights do not give an \textit{exclusive} ``faithful'' explanation, they do provide a meaningful \textit{plausible} explanation. 

This discussion is relevant to our study because it remains unclear whether or not attending to syntactic structure serves, in practice, as plausible explanation for model behavior, or whether or not it is even capable of serving as such. Indeed, the studies of \citet{raganato2018analysis} and \citet{clark-etal-2019-bert} relate a convincing but incomplete picture --- tree decoding accuracy just marginally exceeds baselines and various relations tend to be tracked across varying heads and layers. Thus, our fine-tuning experiments (detailed in the following section) serve to enable an ``easy'' setting wherein we explicitly inform our models of the same structure that we are trying to extract. We posit that, if, after fine-tuning, syntactic structures were still \emph{not} decodable from the attention weights, one could safely conclude that these structures are being stored via a non-transparent mechanism that may not even involve attention weights. Such an insight would allow us to conclude that attention weights cannot provide even a plausible explanation for models relying on syntax. 



\section{Experimental Design}


To examine the extent to which we can decode dependency trees from attention patterns, we run a tree decoding algorithm over
mBERT's attention heads --- before and after fine-tuning via a parsing objective. We surmise that doing so will enable us to determine if attention can be interpreted as a reliable mechanism for capturing linguistic structure. 

\subsection{Model}

We employ mBERT\footnote{\url{https://github.com/google-research/bert}} in our experiments, which has been shown to perform well across a variety of NLP tasks \citep{hu_xtreme_2020,kondratyuk201975} and capture aspects of syntactic structure cross-lingually \citep{pires2019multilingual,chi-etal-2020-finding}. mBERT features 12 layers with 768 hidden units and 12 attention heads, with a joint WordPiece sub-word vocabulary across languages. The model was trained on the concatenation of WikiDumps for the top 104 languages with the largest Wikipedias,
where principled sampling was employed to enforce a balance between high- and low-resource languages. 

\subsection{Decoding Algorithm}
\label{sec:decoding}

For decoding dependency trees, we follow \citet{raganato2018analysis} in applying the Chu-Liu-Edmonds maximum spanning tree algorithm \citep{chu1965shortest} to every layer/head combination available in mBERT ($12 \times 12 = 144$ in total). In order for the matrices to correspond to gold treebank tokenization, we remove the cells corresponding to the BERT delimiter tokens (\texttt{[CLS]} and \texttt{[SEP]}). In addition to this, we sum the columns and average the rows corresponding to the constituent subwords of gold tokens, respectively \citep{clark-etal-2019-bert}. Lastly, since attention patterns across heads may differ in whether they represent heads attending to their dependents or vice versa, we take our input to be the element-wise product of a given attention matrix and its transpose ($A \circ A^\top$). We liken this to the joint probability of a head attending to its dependent and a dependent attending to its head, similarly to \citet{limisiewicz2020universal}. Per this point, we also follow \citet{htut2019attention} in evaluating the decoded trees via Undirected Unlabeled Attachment Score (UUAS) --- the percentage of undirected edges recovered correctly. Since we discount directionality, this is effectively a less strict measure than UAS, but one that has a long tradition in unsupervised dependency parsing since \citet{klein04}. 


\subsection{Data}

For our data, we employ the Parallel Universal Dependencies (PUD) treebanks, as collected in UD v2.4 \citep{11234/1-2988}. PUD was first released as part of the CONLL 2017 shared task \citep{zeman2018conll}, containing 1000 parallel sentences, which were (professionally) translated from English, German, French, Italian, and Spanish to 14 other languages. The sentences are taken from two domains, \textbf{news} and \textbf{wikipedia}, the latter implying some overlap with mBERT's training data (though we did not investigate this). 
We include all PUD treebanks except Thai.\footnote{Thai is the only treebank that does not have a non-PUD treebank available in UD, which we need for our fine-tuning experiments.} 

\subsection{Fine-Tuning Details}
\begin{table*}[h!]
    \def\arraystretch{0.99}
    \footnotesize
    \centering
    \resizebox{\textwidth}{!}{\begin{tabular}{l|cccccccccccccccccc}
        \toprule
        & \textbf{\textsc{ar}}  & \textbf{\textsc{cs}}  & \textbf{\textsc{de}}  & \textbf{\textsc{en}}  & \textbf{\textsc{es}}  & \textbf{\textsc{fi}}  & \textbf{\textsc{fr}}  & \textbf{\textsc{hi}}  & \textbf{\textsc{id}}  & \textbf{\textsc{it}}  & \textbf{\textsc{ja}}  & \textbf{\textsc{ko}}  & \textbf{\textsc{pl}}  & \textbf{\textsc{pt}}  & \textbf{\textsc{ru}}  & \textbf{\textsc{sv}}  & \textbf{\textsc{tr}}  & \textbf{\textsc{zh}} \\
        \midrule
        \textsc{Baseline} & 50 & 40 & 36 & 36 & 40 & 42 & 40 & 46 & 47 & 40 & 43 & 55 & 45 & 41 & 42 & 39 & 52 & 41 \\
        \cmidrule(lr){2-19}
        \multirow{2}{*}{\textsc{Pre}} & 53 & 53 & 49 & 47 & 50 & 48 & 41 & 48 & 50 & 41 & \textbf{45} & 64 & 52 & 50 & 51 & 51 & 55 & 42 \\
         & 7-6 & 10-8 & 10-8 & 10-8 & 9-5 & 10-8 & 2-3 & 2-3 & 9-5 & 6-4 & 2-3 & 9-2 & 10-8 & 9-5 & 10-8 & 10-8 & 3-8 & 2-3 \\
        \midrule
        \multirow{2}{*}{\textsc{None}} & \textbf{76} & \textbf{78} & \textbf{76} & \textbf{71} & \textbf{77} & \textbf{66} & \textbf{45} & \textbf{72} & \textbf{75} & \textbf{58} & 42 & 64 & \textbf{75} & \textbf{76} & \textbf{75} & \textbf{74} & 55 & 38 \\
        & 11-10 & 11-10 & 11-10 & 10-11 & 10-11 & 10-11 & 11-10 & 11-10 & 11-10 & 11-10 & 11-10 & 11-10 & 11-10 & 11-10 & 10-8 & 10-8 & 3-8 & 2-3 \\
        \cmidrule(lr){2-19}
        \multirow{2}{*}{\textsc{Key}} & 62 & 64 & 58 & 53 & 59 & 56 & 41 & 54 & 59 & 47 & 44 & 62 & 64 & 58 & 61 & 59 & 55 & 41\\
        & 10-8 & 10-8 & 11-12 & 10-8 & 11-12 & 10-8 & 7-12 & 10-8 & 10-8 & 9-2 & 2-3 & 10-8 & 10-8 & 11-12 & 10-8 & 12-10 & 3-12 & 2-3\\
        \cmidrule(lr){2-19}
        \multirow{2}{*}{\textsc{Query}} & 69 & 74 & 70 & 66 & 73 & 63 & 42 & 62 & 67 & 54 & \textbf{45} & 65 & 72 & 70 & 70 & 68 & 56 & 42 \\
        & 11-4 & 10-8 & 11-4 & 11-4 & 11-4 & 10-8 & 11-4 & 11-4 & 11-4 & 11-4 & 2-3 & 10-8 & 11-4 & 11-4 & 10-8 & 11-4 & 10-8 & 2-3\\
        \cmidrule(lr){2-19}
        \multirow{2}{*}{\textsc{KQ}} & 71 & 76 & 70 & 65 & 74 & 62 & 43 & 64 & 69 & 55 & 44 & 64 & 73 & 73 & 69 & 69 & 55 & 41 \\
        & 11-4 & 11-4 & 11-4 & 11-4 & 11-4 & 11-4 & 10-11 & 11-4 & 11-4 & 11-4 & 2-3 & 11-4 & 11-4 & 11-4 & 11-4 & 11-4 & 11-4 & 2-3 \\
        \cmidrule(lr){2-19}
        \multirow{2}{*}{\textsc{Value}} & 75 & 72 & 72 & 64 & 76 & 59 & \textbf{45} & 63 & 73 & 55 & \textbf{45} & \textbf{66} & 73 & 74 & 69 & 65 & \textbf{57} & \textbf{42} \\
        & 12-5 & 12-5 & 12-5 & 12-5 & 12-5 & 12-5 & 12-5 & 12-5 & 12-5 & 12-5 & 2-3 & 10-8 & 12-5 & 12-5 & 12-5 & 12-5 & 12-5 & 3-8 \\
        \cmidrule(lr){2-19}
        \multirow{2}{*}{\textsc{Dense}} & 68 & 71 & 65 & 60 & 67 & 61 & 42 & 65 & 66 & 49 & 44 & 64 & 70 & 64 & 67 & 64 & 55 & 40 \\
        & 11-10 & 11-10 & 11-10 & 10-8 & 12-10 & 11-10 & 10-8 & 11-10 & 11-10 & 9-5 & 3-12 & 11-10 & 11-10 & 12-5 & 11-10 & 11-10 & 11-10 & 3-12 \\
        \bottomrule
    \end{tabular}}
    \caption{Adjacent-branching baseline and maximum UUAS decoding accuracy per PUD treebank, expressed as best score and best layer/head combination for UUAS decoding. \textsc{Pre} refers to basic mBERT model before fine-tuning, while all cells below correspond different fine-tuned models described in Section 3.4. Best score indicated in \textbf{bold}.}
    \label{tab:full_results}
\end{table*}

In addition to exploring pretrained mBERT's attention weights, we are also interested in how attention might be guided by a training objective that learns the exact tree structure we aim to decode. To this end, we employ the graph-based decoding algorithm of the biaffine parser introduced by \citet{dozat2016deep}. We replace the standard BiLSTM encoder for this parser 
with the entire mBERT network, which we fine-tune with the parsing loss. The full parser decoder consists of four dense layers, two for head/child representations for dependency arcs (dim. 500) and two for head/child representations for dependency labels (dim. 100). These are transformed into the label space via a bilinear transform.

After training the parser, we can decode the fine-tuned mBERT parameters in the same fashion as described in Section~\ref{sec:decoding}. We surmise that, if attention heads are capable of tracking hierarchical relations between words in any capacity, it is precisely in this setting that this ability would be attested. In addition to this, we are interested in what individual \textit{components} of the mBERT network are capable of steering attention patterns towards syntactic structure. We believe that addressing this question will help us not only in interpreting decisions made by BERT-based neural parsers, but also in aiding us developing syntax-aware models in general \citep{strubell-etal-2018-linguistically,swayamdipta-etal-2018-syntactic}. As such --- beyond fine-tuning all parameters of the mBERT network (our basic setting) --- we perform a series of ablation experiments wherein we update only one set of parameters per training cycle, e.g. the Query weights $W_{i}^{Q}$, and leave everything else frozen. This gives us a set of 6 models, which are described below. For each model, all non-BERT parser components are always left unfrozen.

\begin{itemize}[itemsep=0pt,topsep=3pt]
\setlength{\parskip}{.1mm}
\setlength{\parsep}{0.1mm}
    \item \textsc{Key}: only the $K$ components of the transformer are unfrozen; these are the representations of tokens that are paying attention \textit{to} other tokens.
    \item \textsc{Query}: only the $Q$ components are unfrozen; these, conversely, are the representations of tokens being paid attention to.
    \item \textsc{KQ}: both keys and queries are unfrozen.
    \item \textsc{Value}: semantic value vectors per token ($V$) are unfrozen; they are composed after being weighted with attention scores obtained from the $K$/$Q$ matrices.
    \item \textsc{Dense}: the dense feed-forward networks in the attention mechanism; all three per layer are unfrozen.
    \item \textsc{None}: The basic setting with nothing frozen; all parameters are updated with the parsing loss. 
\end{itemize}

\noindent
We fine-tune each of these models on a concatentation of all PUD treebanks for 20 epochs, which effectively makes our model multilingual. We do so in order to 1) control for domain and annotation confounds, since all PUD sentences are parallel and are natively annotated (unlike converted UD treebanks, for instance); 2) increase the number of training samples for fine-tuning, as each PUD treebank features only 1000 sentences; and 3) induce a better parser through multilinguality, as in \citet{kondratyuk-straka-2019-75}. Furthermore, in order to gauge the overall performance of our parser across all ablated settings, we evaluate on the test set of the largest non-PUD treebank available for each language, since PUD only features test partitions. When training, we employ a combined dense/sparse Adam optimiser, at a learning rate of $3 * 10^{-5}$. We rescale gradients to have a maximum norm of 5.

\section{Decoding mBERT Attention}

\begin{figure}[h!]
        \centering
        \includegraphics[scale=0.15]{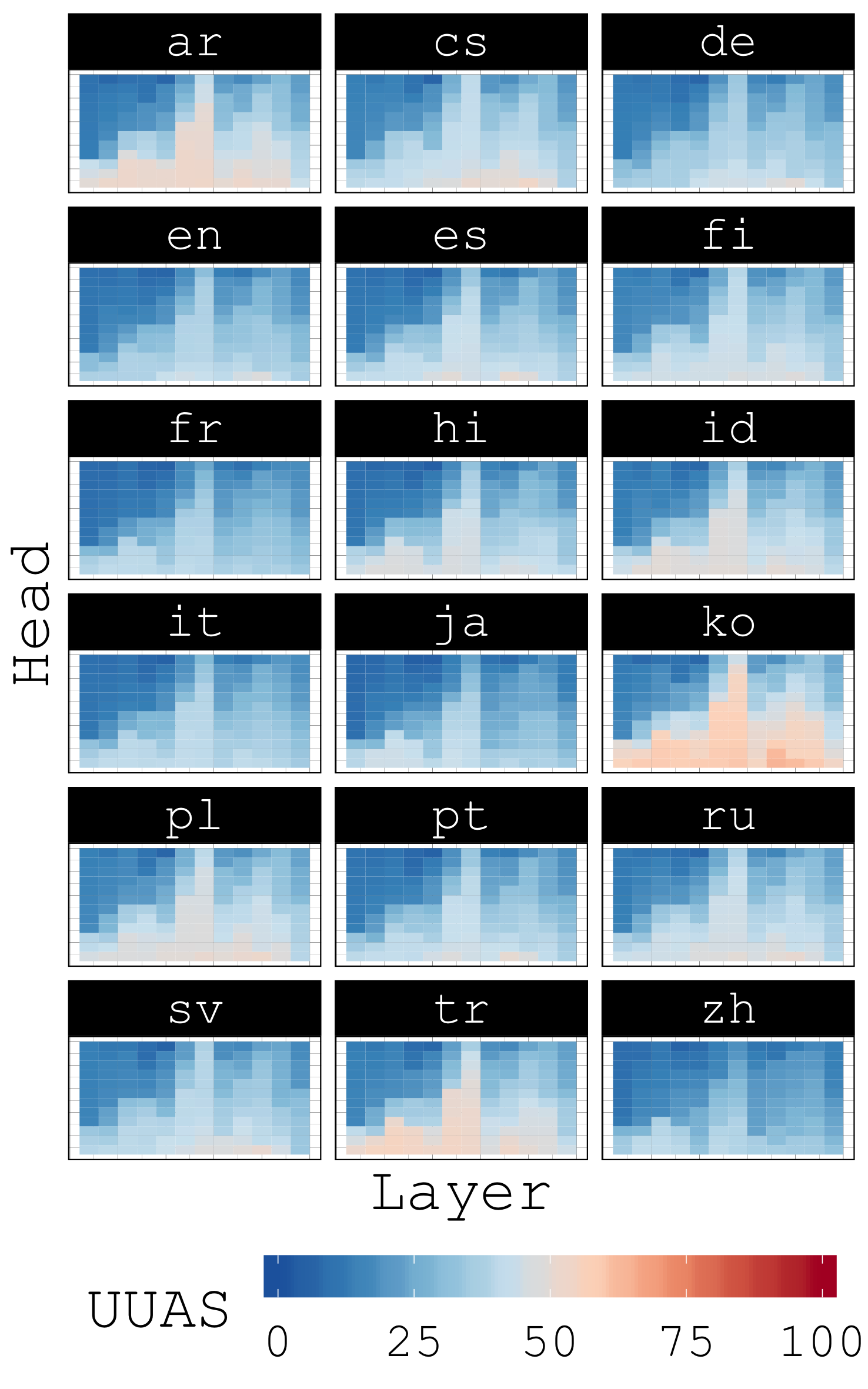}
        \caption{UUAS of MST decoding per layer and head, across languages. Heads (y-axis) are sorted by accuracy for easier visualization.}
        \label{fig:raw_uuas_results}
\end{figure}

\begin{figure*}[h]
    \centering
    \includegraphics[scale=0.11]{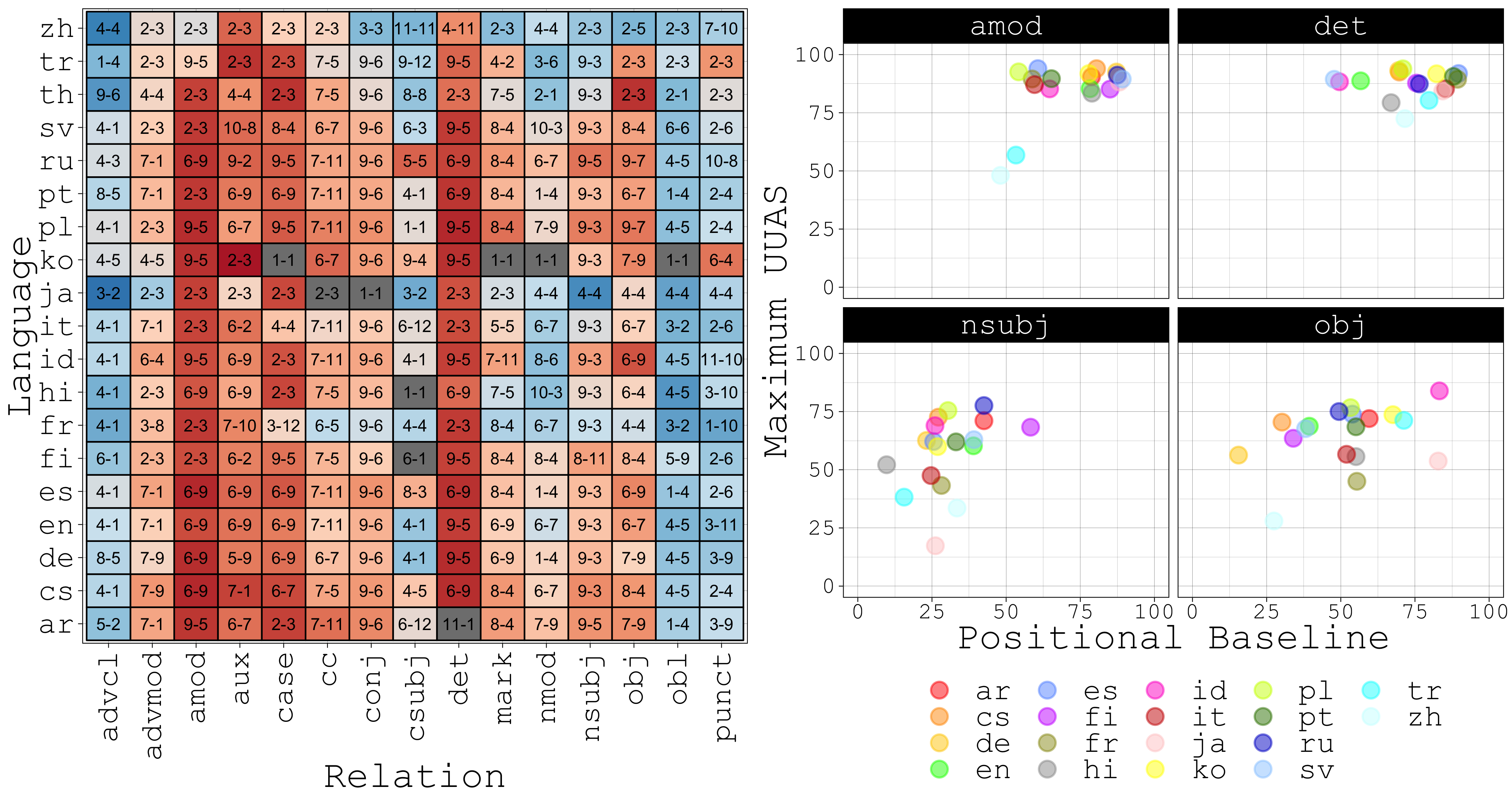}
    \caption{Left: UUAS per relation across languages (best layer/head combination indicated in cell). Right: Best UUAS as a function of best positional baseline (derived from the treebank), selected relations.}
    \label{fig:raw_relations}
\end{figure*}

The second row of Table \ref{tab:full_results} (\textsc{Pre}) depicts the UUAS after running our decoding algorithm over mBERT attention matrices, per language. We see a familiar pattern to that in \citet{clark-etal-2019-bert} among others --- namely that attention patterns extracted directly from mBERT appear to be incapable of decoding dependency trees beyond a threshold of 50--60\% UUAS accuracy.
However, we also note that, in all languages, the attention-decoding algorithm outperforms a \textsc{Baseline} (row 1) that draws an (undirected) edge between any two adjacent words in linear order, 
which implies that some non-linear structures are captured with regularity. Indeed, head 8 in layer 10 appears to be particularly strong in this regard, returning the highest UUAS for 7 languages. Interestingly, the accuracy patterns across layers depicted in Figure~\ref{fig:raw_uuas_results} tend to follow an identical trend for all languages, with nearly all heads in layer 7 returning high within-language accuracies. 

It appears that attention for some languages (Arabic, Czech, Korean, Turkish) is comparatively easier to decode than others (French, Italian, Japanese, Chinese). A possible explanation for this result is that dependency relations between content words, which are favored by the UD annotation, are more likely to be adjacent in the morphologically rich languages of the first group (without intervening function words). This assumption seems to be corroborated by the high baseline scores for Arabic, Korean and Turkish (but not Czech). Conversely, the low baselines scores and the likewise low decoding accuracies for the latter four languages are difficult to characterize. Indeed, we could not identify what factors --- typological, annotation, tokenization or otherwise --- would set French and Italian apart from the remaining languages in terms of score. However, we hypothesize that the tokenization and our treatment of subword tokens plays a part in attempting to decode attention from Chinese and Japanese representations. Per the mBERT documentation,\footnote{\url{https://github.com/google-research/bert/blob/master/multilingual.md}} Chinese and Japanese Kanji character spans within the CJK Unicode range are character-tokenized. This lies in contrast with all other languages (Korean Hangul and Japanese Hiragana and Katakana included), which rely on whitespace and WordPiece \citep{wu2016google}. It is thus possible that the attention distributions for these two languages (at least where CJK characters are relevant) are devoted to composing words, rather than structural relations, which will distort the attention matrices that we compute to correspond with gold tokenization (e.g. by maxing rows and averaging columns).   


\paragraph{Relation analysis} We can disambiguate what sort of structures are captured with regularity by looking at the UUAS returned per dependency relation. Figure \ref{fig:raw_relations} (left) shows that adjectival modifiers (\texttt{amod}, mean UUAS = $85$  $\pm 12$) 
  and determiners (\texttt{det}, $88\pm6$) are among the easiest relations to decode across languages. Indeed, words that are connected by these relations are often adjacent to each other and may be simple to decode if a head is primarily concerned with tracking linear order. To verify the extent to which this might be happening, we plot the aforementioned decoding accuracy as a function of select relations' positional baselines in Figure \ref{fig:raw_relations} (right). The positional baselines, in this case, are calculated by picking the most frequent offset at which a dependent occurs with respect to its head, e.g., $-$1 for \texttt{det} in English, meaning one position to the left of the head. Interestingly, while we observe significant variation across the positional baselines for \texttt{amod} 
  and \texttt{det}, the decoding accuracy remains quite high.

In slight contrast to this, the core subject (\texttt{nsubj}, $58\pm16$ SD) and object (\texttt{obj}, $64\pm13$) relations prove to be more difficult to decode. Unlike the aforementioned relations, \texttt{nsubj} and \texttt{obj} are much more sensitive to the word order properties of the language at hand. For example, while a language like English, with Subject-Verb-Object (SVO) order, might have the subject frequently appear to the left of the verb, an SOV language like Hindi might have it several positions further away, with an object and its potential modifiers intervening. Indeed, the best positional baseline for English \texttt{nsubj} is 39 UUAS, while it is only 10 for Hindi. Despite this variation, the relation seems to be tracked with some regularity by the same head (layer 3, head 9), returning 60 UUAS for English and 52 for Hindi. The same can largely be said for \texttt{obj}, where the positional baselines return $51\pm18$. In this latter case, however, the heads tend to be much differently distributed across languages. Finally, he results for the \texttt{obj} relation provides some support for our earlier explanation concerning morphologically rich languages, as Arabic, Czech, Korean and Turkish all have among the highest accuracies (as well as positional baselines).

\section{Fine-Tuning Experiments}
\label{sec:ft}
\begin{figure*}[ht!]
    \centering
    \includegraphics[width=\textwidth]{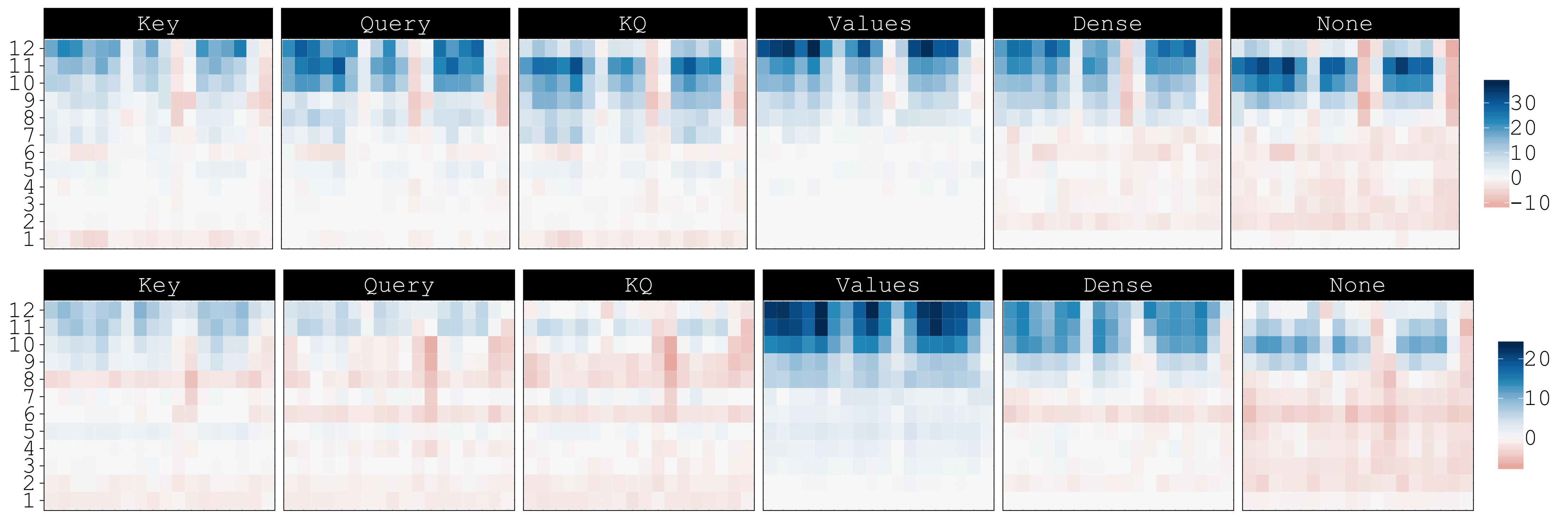}
    \caption{(Top) best scores across all heads, per language; (bottom) mean scores across all heads, per language. The languages (hidden from the X-axis for brevity) are, in order, \emph{ar, cs, de, en, es, fi, fr, hi, id, it, ja, ko, pl, pt, ru, sv, tr, zh}}
    \label{fig:all_models_max}
\end{figure*}

\begin{figure}[ht!]
    \centering
    \includegraphics[scale=0.10]{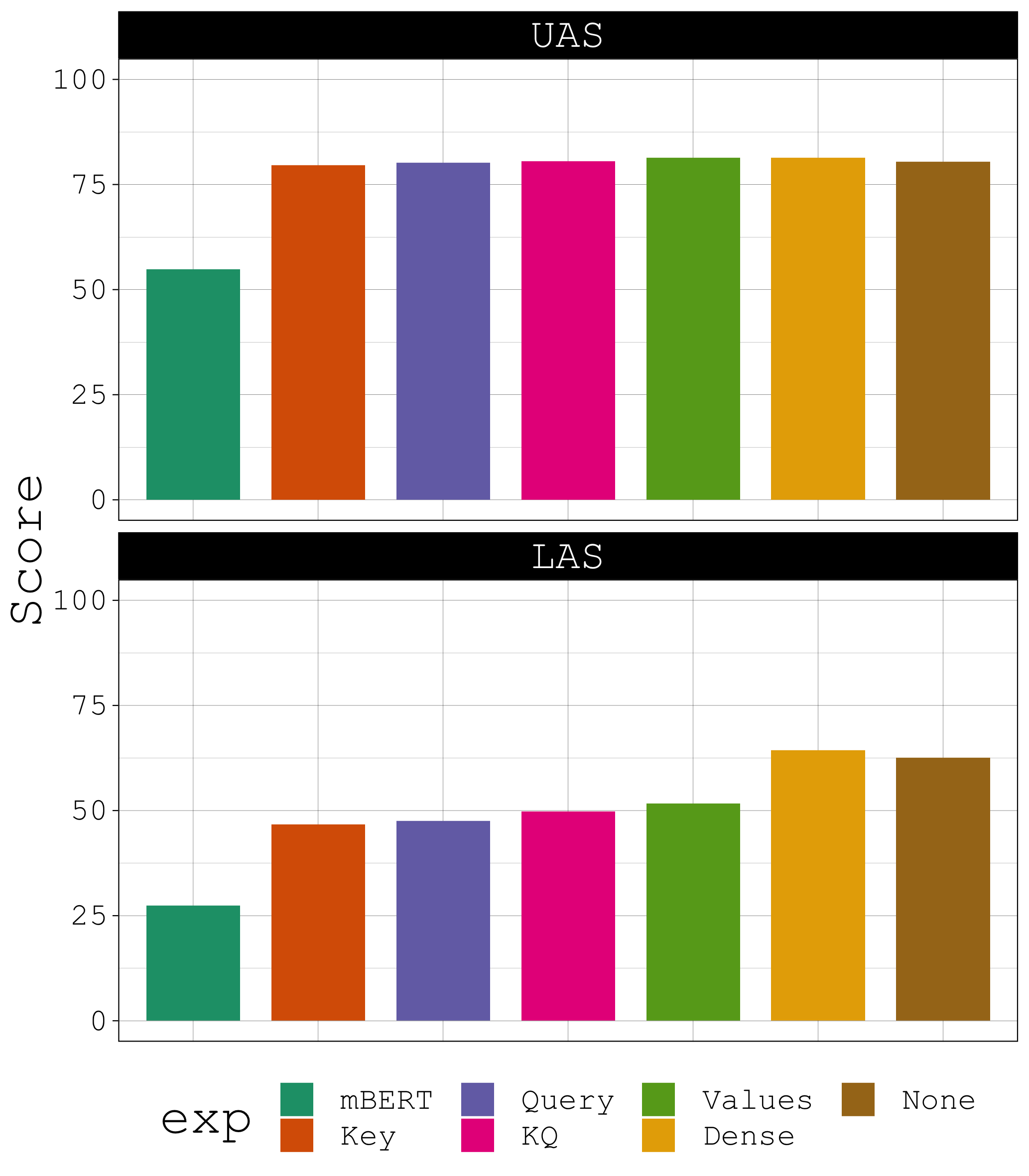}
    \caption{Mean UAS and LAS when evaluating different models on language-specific treebanks (Korean excluded due to annotation differences). \textsc{mBERT} refers to models where the entire mBERT network is frozen as input to the parser.}
    \label{fig:supervised}
\end{figure}
Next, we investigate the effect fine-tuning has on UUAS decoding. Row 3 in Table~\ref{tab:full_results} (\textsc{None}) indicates that fine-tuning does result in large improvements to UUAS decoding across most languages, often by margins as high as $\sim30\%$. This shows that with an explicit parsing objective, attention heads are capable of serving as explanatory mechanisms for syntax; syntactic structure can be made to be transparently stored in the heads, in a manner that does not require additional probe fitting or parameterized transformation to extract.

Given that we do manage to decode reasonable syntactic trees, we can then refine our question --- what components are capable of learning these trees? One obvious candidate is the key/query component pair, 
given that attention weights are a scaled softmax of a composition of the two. 
Figure \ref{fig:all_models_max} (top) shows the difference between pretrained UUAS and fine-tuned UUAS per layer, across models and languages. Interestingly, the best parsing accuracies do not appear to vary much depending on what component is frozen. We do see a clear trend, however, in that decoding the attention patterns of the fine-tuned model typically yields better UUAS than the pretrained model, particularly in the highest layers. Indeed, the lowest layer at which fine-tuning appears to improve decoding is layer 7. This implies that, regardless of which component remains frozen, the parameters facing any sort of significant and positive update tend to be those appearing towards the higher-end of the network, closer to the output. 



For the frozen components, the best improvements in UUAS are seen at the final layer in \textsc{Value}, which is also the only model that shows consistent improvement, as well as the highest average improvement in mean scores\footnote{The inner average is over all heads; the outer is over all languages.} for the last few layers. 
Perhaps most interestingly, the mean UUAS (Figure \ref{fig:all_models_max} (bottom)) for our ``attentive'' components -- keys, queries, and their combination -- does not appear to have improved by much after fine-tuning. In contrast, the maximum does show considerable improvement; this seems to imply that although all components appear to be more or less equally capable of learning decodable heads, the attentive components, when fine-tuned, appear to sharpen fewer heads. 

Note that the only difference between keys and queries in an attention mechanism is that keys are transposed to index attention from/to appropriately. Surprisingly, \textsc{Key} and \textsc{Query} appear to act somewhat differently, with \textsc{Query} being almost uniformly better than \textsc{Key} with the best heads, whilst \textsc{Key} is slightly better with averages, implying distinctions in how both store information. Furthermore, allowing both keys and queries seems to result in an interesting contradiction -- the ultimate layer, which has reasonable maximums and averages for both \textsc{Key} and \textsc{Query}, now seems to show a UUAS drop almost uniformly. This is also true for the completely unfrozen encoder. 

\paragraph{Supervised Parsing} In addition to decoding trees from attention matrices, we also measure supervised UAS/LAS on a held-out test set.\footnote{Note that the test set in our scenario is from the actual, non-parallel language treebank; as such, we left Korean out of this comparison due to annotation differences.} Based on Figure~\ref{fig:supervised}, it is apparent that all settings result in generally the same UAS. This is somewhat expected; \citet{lauscher_zero_2020} see better results on parsing with the entire encoder frozen, implying that the task is easy enough for a biaffine parser to learn, given frozen mBERT representations.\footnote{Due to training on concatenated PUD sets, however, our results are not directly comparable/} The LAS distinction is, however, rather interesting: there is a marked difference between how important the dense layers are, as opposed to the attentive components. This is likely not reflected in our UUAS probe as, strictly speaking, labelling arcs is not equivalent to searching for structure in sentences, but more akin to classifying pre-identified structures. We also note that \textsc{Dense} appears to be better than \textsc{None} on average, implying that non-dense components might actually be hurting labelling capacity.

In brief, consolidating the two sets of results above, we can draw three interesting conclusions about the components:

\begin{enumerate}
\setlength{\parskip}{.1mm}
\setlength{\parsep}{0.1mm}
\setlength{\itemsep}{0.3mm}

    \item \textbf{Value} vectors are best aligned with syntactic dependencies; this is reflected both in the best head at the upper layers, and the average score across all heads.
    \item \textbf{Dense} layers appear to have moderate informative capacity, but appear to have the best learning capacity for the task of arc labelling.
    \item Perhaps most surprisingly, \textbf{Key} and \textbf{Query} vectors do not appear to make any outstanding contributions, save for sharpening a smaller subset of heads.
\end{enumerate}

\noindent
Our last result is especially surprising for UUAS decoding. Keys and queries, fundamentally, combine to form the attention weight matrix, which is precisely what we use to decode trees. One would expect that allowing these components to learn from labelled syntax would result in the best improvements to decoding, but all three have surprisingly negligible mean improvements. This indicates that we need to further improve our understanding of how attentive structure and weighting really works. 

\paragraph{Cross-linguistic observations} We notice no clear cross-linguistic trends here across different component sets; however, certain languages do stand out as being particularly hard to decode from the fine-tuned parser. These include Japanese, Korean, Chinese, French and Turkish. For the first three, we hypothesise that tokenization clashes with mBERT's internal representations may play a role. Indeed, as we hypothesized in Section \ref{sec:decoding}, it could be the case that the composition of CJK characters into gold tokens for Chinese and Japanese may degrade the representations (and their corresponding attention) therein. Furthermore, for Japanese and Korean specifically, it has been observed that tokenization strategies employed by different treebanks could drastically influence the conclusions one may draw about their inherent hierarchical structure \citep{kulmizev-etal-2020-neural}. Turkish and French are admittedly more difficult to diagnose. Note, however, that we fine-tuned our model on a concatenation of all PUD treebanks. As such, any deviation from PUD's annotation norms is therefore likely to be heavily penalised, by virtue of signal from other languages drowning out these differences.

\section{Conclusion}

In this study, we revisited the prospect of decoding dependency trees from the self-attention patterns of Transformer-based language models. We elected to extend our experiments to 18 languages in order to gain better insight about how tree decoding accuracy might be affected in the face of (modest) typological diversity. Surprisingly, across all languages, we were able to decode dependency trees from attention patterns more accurately than 
an adjacent-linking baseline, implying that some structure was indeed being tracked by the mechanism. In looking at specific relation types, we corroborated previous studies in showing that particular layer-head combinations tracked the same relation with regularity across languages, despite typological differences concerning word order, etc. 

In investigating the extent to which attention can be guided to properly capture structural relations between input words, we fine-tuned mBERT as input to a dependency parser. This, we found, yielded large improvements over the pretrained attention patterns in terms of decoding accuracy, demonstrating that the attention mechanism was learning to represent the structural objective of the parser. In addition to fine-tuning the entire mBERT network, we conducted a series of experiments, wherein we updated only select components of model and left the remainder frozen. Most surprisingly, we observed that the Transformer parameters designed for composing the attention matrix, $K$ and $Q$, were only modestly capable of guiding the attention towards resembling the dependency structure. In contrast, it was the Value ($V$) parameters, which are used for computing a weighted sum over the $KQ$-produced attention, that yielded the most faithful representations of the linguistic structure via attention. 

Though prior work~\citep{kovaleva_revealing_2019,zhao_attention_2020} seems to indicate that there is a lack of a substantial change in attention patterns after fine-tuning on syntax- and semantics-oriented classification tasks, the opposite effect has been observed with fine-tuning on negation scope resolution, where a more explanatory attention mechanism can be induced~\citep{htut2019attention}. Our results are similar to the latter, and we demonstrate that given explicit syntactic annotation, attention weights do end up storing more transparently decodable structure. It is, however, still unclear which sets of transformer parameters are best suited for learning this information and storing it in the form of attention. 

\section*{Acknowledgements}
Our experiments were run on resources provided by UNINETT Sigma2 - the National Infrastructure for High Performance Computing and 
Data Storage in Norway, under the NeIC-NLPL umbrella. Mostafa and Anders were funded by a Google Focused Research Award. We would like to thank Daniel Dakota and Ali Basirat for some fruitful discussions and the anonymous reviewers for their excellent feedback.

\bibliography{eacl2021}
\bibliographystyle{acl_natbib}

\newpage
\appendix

\begin{figure*}
    \centering
    \includegraphics[scale=0.05]{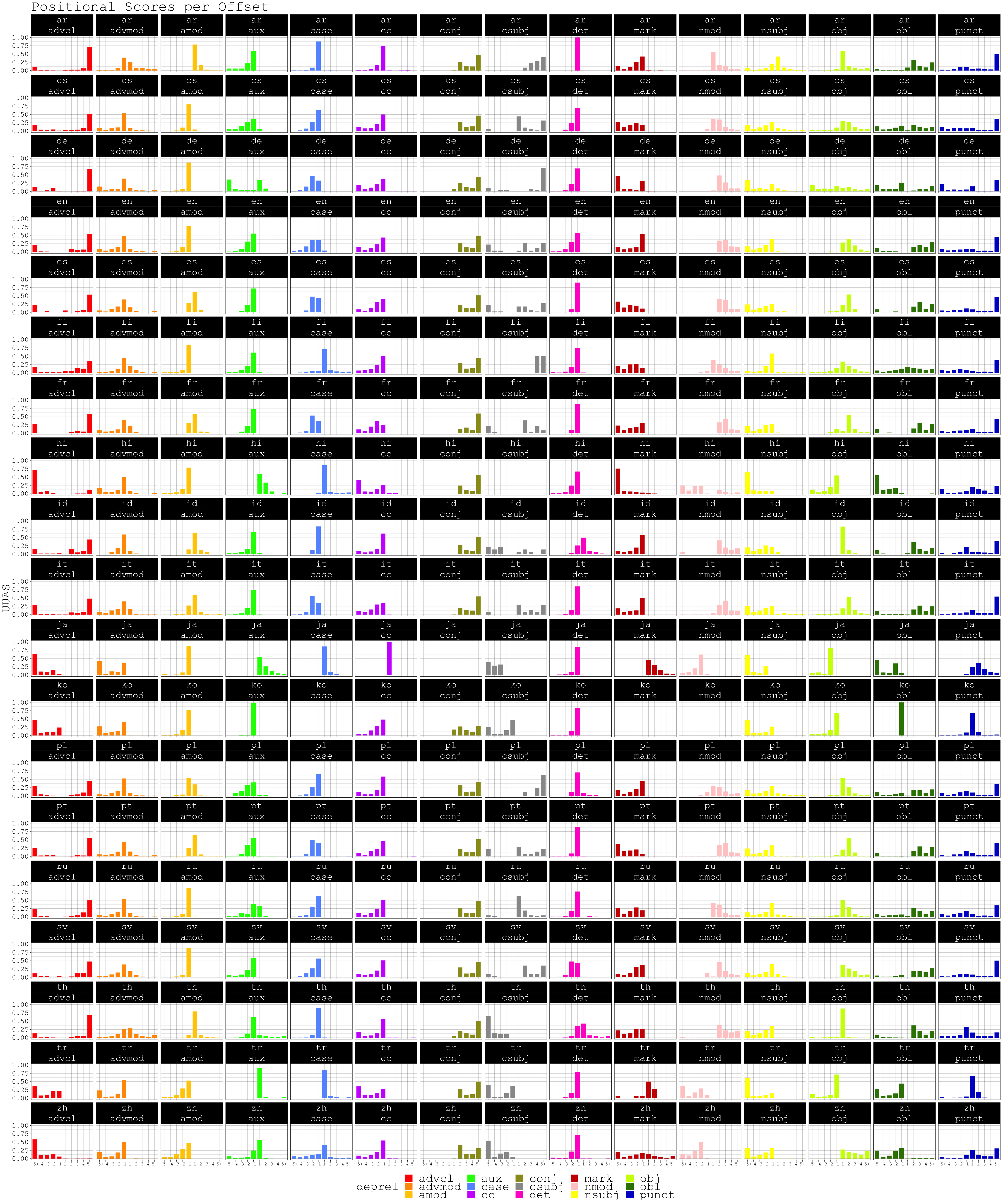}
    \caption{Positional scores across relations for all languages.}
    \label{fig:my_label}
\end{figure*}

\begin{figure*}
    \centering
    \includegraphics[width=\textwidth]{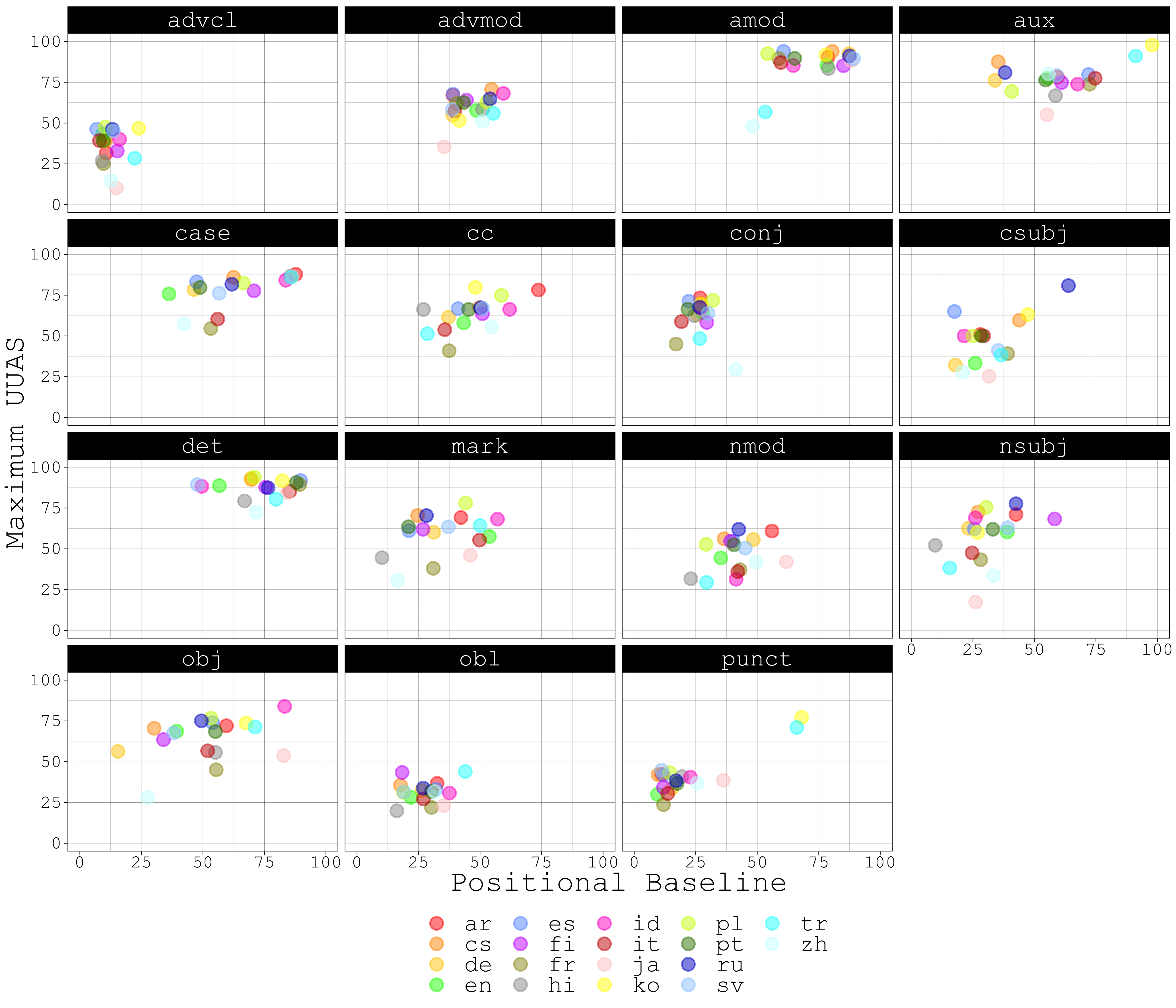}
    \caption{Decoding UUAS as a function of best positional baselines.}
    \label{fig:my_label}
\end{figure*}

\begin{figure*}
    \centering
    \includegraphics[width=\textwidth]{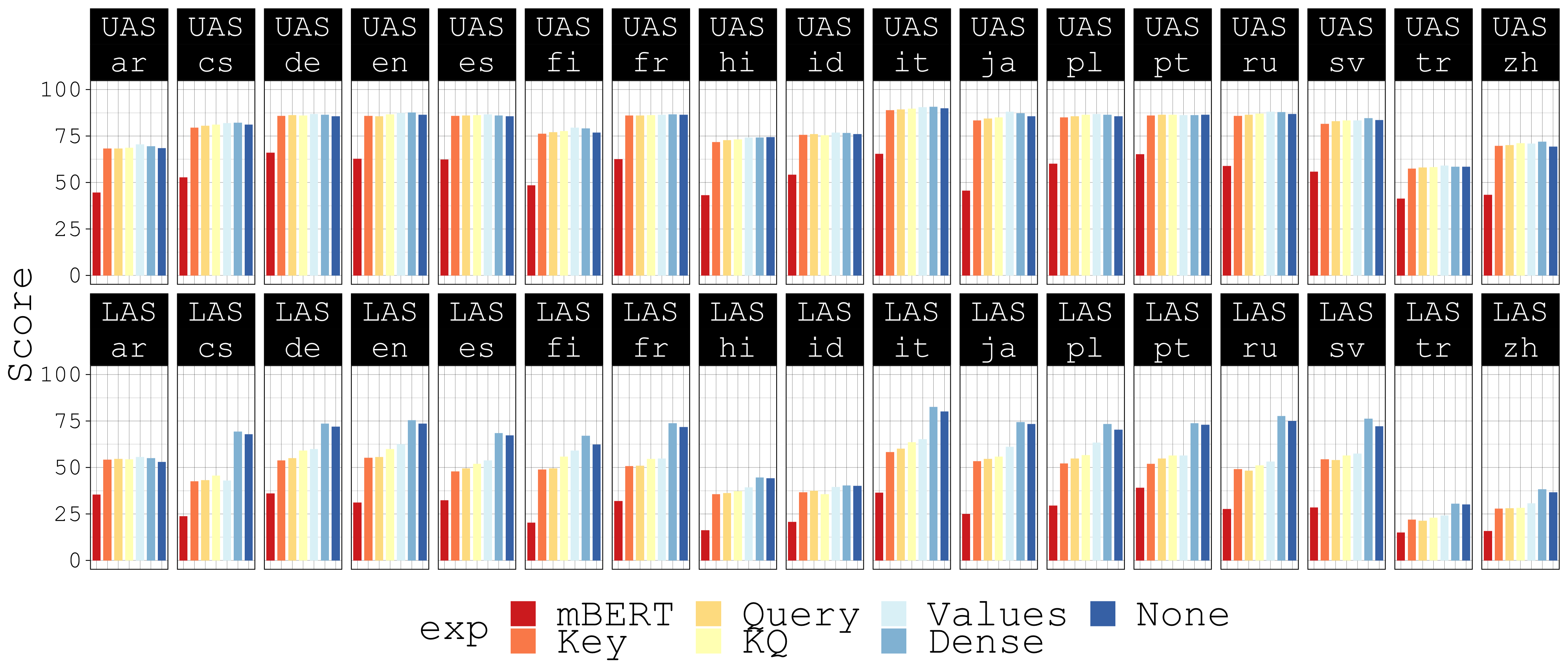}
    \caption{Parsing scores across components and languages.}
    \label{fig:my_label}
\end{figure*}

\end{document}